\documentclass{article}
\usepackage{multicol}

\usepackage[preprint, nonatbib]{neurips_2021}

\usepackage[utf8]{inputenc} 
\usepackage[T1]{fontenc}    
\usepackage{hyperref}       
\usepackage{url}            
\usepackage{booktabs}       
\usepackage{amsfonts}       
\usepackage{nicefrac}       
\usepackage{microtype}      
\usepackage{xcolor}         
\usepackage{graphicx}
\usepackage[style=ieee]{biblatex}
\title{Minimizing the Effect of Noise and Limited Dataset Size in Image Classification Using Depth Estimation as an Auxiliary Task with Deep Multitask Learning}

\author{
  Khashayar Namdar\textsuperscript{1,2,6}\thanks{KN and PV are shared first authors}\\
  \And
  Partoo Vafaeikia\textsuperscript{1,2}* \\
  \AND
  Farzad Khalvati \textsuperscript{1,2,3,4,5,6} \\
}

\begin{document}

\maketitle

\newcommand\imgdim{40mm}
\

{\footnotesize \textsuperscript{1}Institute of Medical Science, University of Toronto, Toronto, ON, Canada\\
\textsuperscript{2}Department of Diagnostic Imaging, Research institute, The Hospital for Sick Children, Toronto, ON, Canada\\
\textsuperscript{3}Department of Medical Imaging, University of Toronto, Toronto, ON\\
\textsuperscript{4}Department of Computer Science, University of Toronto, Toronto, ON, Canada\\
\textsuperscript{5}Department of Mechanical and Industrial Engineering, University of Toronto\\
\textsuperscript{6}Vector Institute, Toronto, ON, Canada\\}

\begin{abstract}
  Generalizability is the ultimate goal of Machine Learning (ML) image classifiers, for which noise and limited dataset size are among the major concerns. We tackle these challenges through utilizing the framework of deep Multitask Learning (dMTL) and incorporating image depth estimation as an auxiliary task. On a customized and depth-augmented derivation of the MNIST dataset, we show a) multitask loss functions are the most effective approach of implementing dMTL, b) limited dataset size primarily contributes to classification inaccuracy, and c) depth estimation is mostly impacted by noise. In order to further validate the results, we manually labeled the NYU Depth V2 dataset for scene classification tasks. As a contribution to the field, we have made the data in python native format publicly available as an open-source dataset and provided the scene labels. Our experiments on MNIST and NYU-Depth-V2 show dMTL improves generalizability of the classifiers when the dataset is noisy and the number of examples is limited.

\end{abstract}
\section{Introduction}
Machine Learning (ML) models, specifically Convolutional Neural Networks (CNN), are proven to be promising in different fields and contexts such as medical imaging, gaming, real estate, and transportation. However, their performance and generalization are conditioned on their access to a large amount of standardized and pre-processed annotated datasets. Aggregating labeled, clean, and abundant data samples is not a straightforward task. Hence, our goal is to decrease the sensitivity of the performance of CNNs to dataset size and noise in the context of depth estimation. In this study, we create multiple custom datasets based on MNIST~\cite{LeCun2010MNIST} and NYU-Depth-V2 ~\cite{Silberman:ECCV12} and utilize deep Multitask Learning (dMTL) to perform classification and depth estimation tasks.

\subsection{Multitask Learning}

The concept of dMTL is to perform multiple tasks simultaneously. The method of sharing parameters in dMTL, which is either hard \cite{Baxter1997Bayesian/Information} or soft \cite{Duong2015Low}, determines how the model is optimized for multiple learning tasks at hand \cite{Girshick2015Fast}. Even if the ultimate goal is to make predictions for a single task, dMTL can significantly improve the performance of the specific task if the main task is augmented with other tasks (i.e., auxiliary tasks) \cite{Baxter1997Bayesian/Information}. dMTL leads to a better generalization of the model through introducing an inductive bias based on the comprehensive information in the image \cite{Ruder2017Overview}.

Considering our scenario and building upon the prior research on multitask learning, we regard depth estimation and classification as entangled tasks. We examine the effect of noise and dataset size on these tasks and form a dMTL setting to improve their performance effectively.

\subsection{Depth Estimation Methods}

There are two main approaches for depth estimation; unsupervised and supervised. In unsupervised methods, stereo images are utilized to infer the 3D structure of the scene. However, in supervised methods, which is our area of interest, the model is trained on existing pixel-level ground truth depth maps. Some methods use partial depth maps in training phase, and as a result, different requirements and constraints need to be introduced to achieve better performance.

Authors of "Depth Map Prediction from a Single Image using a Multi-Scale Deep Network \cite{Eigen2014Depth}" obtained ground truth depth information of RGB images from Kinect or laser data for training, however, during inference, depth is estimated only using the RGB images. The main idea of this method is to achieve better performance by utilizing information obtained from both global and local views. The neural network contains two components: one for estimating the global structure of the scene, and a second that refines the first one by utilizing local information. The loss function explicitly accounts for depth relations between pixel locations as well as the pointwise error. This approach has been tested on NYU Depth and KITTI \cite{Geiger2012Are} datasets and has achieved better performance compared to Make3D \cite{Saxena2007Learning}, Karsch et al \cite{Karsch2014Depth} and Ladicky et al. \cite{Ladický2014Pulling} methods. 

"Deep Ordinal Regression Network for Monocular Depth Estimation \cite{Fu2018Deep}" phrases the problem as an ordinal regression task instead of a continuous depth map prediction, where original depth maps are discretized into a number of intervals using a space-increasing discretization (SID) strategy. Although one approach is to pass discretized depth values to a multi-class classifier, but this strategy would lose the importance of ordering. Therefore, the problem is casted as an ordinal regression problem. The network architecture is divided into two sections: a feature extractor and a scene understanding module to capture global contextual information. Training a regression network is slow in convergence and might result in underfitting. Using ordinal regression method instead, elevates the overall performance.

\section{Related Work}
Several advancements have been made in utilizing dMTL with depth datasets, some of which focus on optimizing classification tasks and have been influential in our research.

\begin{figure}[htbp]
  \centering
  \includegraphics[scale=0.55]{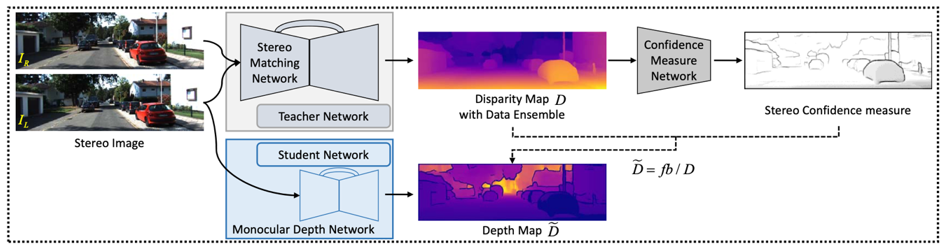}
  \caption{First step is to generate depth maps from stereo image pairs using deep teacher network trained with ground truth depth data. The stereo confidence map is also estimated to recognize inaccurate depth values. Furthermore, the monocular student network is trained with the output of the teacher network, pseudo ground truth depth maps and stereo confidence maps. Reprinted from \cite{Cho2019Large}}
  \label{figure:1}
\end{figure}

\textbf{A Large RGB-D Dataset for Semi-supervised Monocular Depth Estimation \cite{Cho2019Large}} utilizes a semi-supervised approach based on the student-teacher strategy \cite{Xu2019Training}. The teacher network uses a massive unlabeled stereo image dataset as an input. Using off-the-shelf stereo matching CNN \cite{Pang2017Cascade} trained with ground truth depth data, depth maps of the images are created. To improve performance, stereo confidence maps \cite{Park2015Leveraging} are used as an auxiliary data. These confidence maps are also used in the student network to avoid utilization of inaccurate stereo depth values. Finally, outputted depth maps and stereo confidence maps are used as pseudo ground truth to supervise the monocular student network. The overall framework is illustrated in Figure~\ref{figure:1}.

\textbf{MultiDepth: Single-Image Depth Estimation via Multi-Task Regression and Classification \cite{Liebel2019MultiDepth:}} takes the novel MTL approach to monocular depth estimation to overcome the instability and slow convergence of depth estimation during supervised training. It uses depth interval classification as an auxiliary task during training phase and at inference, it only predicts continuous depth values by disabling the use of that auxiliary task. This approach has been tested on the KITTI dataset for road scenes. The ground truth depth information obtained from LiDAR sensors are used as the training data for the main regression task. They are also classified into a number of intervals to be used as the auxiliary task and improve the performance of single image depth estimation.

\textbf{Estimated Depth Map Helps Image Classification\cite{He2017Estimated}} investigates the effect of including estimated depth maps as an extra channel to the network to demonstrate improvements in image classification task. To achieve this, depth maps for the CIFAR-10 RGB images are estimated using a trained deep convolutional neural field model. Next, depth maps are added as a new channel along with RGB images to build a RGBD dataset. Finally, the newly created RGBD dataset is used as an input to the image classifier to detect which category they belong to out of the original 10 classes. This technique, has improved the accuracy by 0.55\% in ResNet-20.

\section{Dataset}
Classification is a popular topic in computer vision where the aim is to either classify images based on their overall theme or classify objects in images to improve image understanding and analysis. There are numerous datasets which are formed for the task of classification. In contrast, depth estimation demands complex ground truths of depth maps, and thus dataset options are limited. We worked on two different paradigms: customized MNIST-based datasets and manually classified NYU Depth V2.

\subsection{MNIST}
\begin{figure}[htbp]
  \centering
  \includegraphics[scale=0.45]{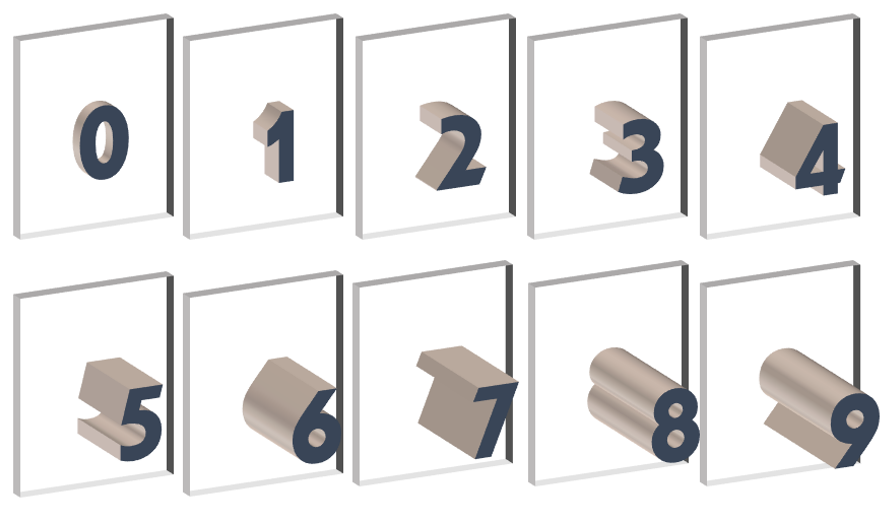}
  \caption{Illustration of our extruded MNIST}
  \label{figure:2}
\end{figure}

MNIST \cite{LeCun2010MNIST} is a dataset of handwritten digits, commonly used in classification problems. MNIST exists in multiple formats, such as Numpy, Pickle, and even in the form of tabular data, and is included in different libraries (e.g., torchvision\footnote{https://pytorch.org/vision/stable/datasets.html} and tensorflow datasets\footnote{https://www.tensorflow.org/datasets/catalog/mnist}). In this research a series of customized MNIST based datasets is created by introducing noise to specific images and resizing them. Originally, MNIST images have the fixed size of 28x28. There are 60,000 and 10,000 images in training and test cohorts, respectively. To approach real-world images, and to make the trained models suitable for Transfer Learning (TL) purposes, each image is upsampled to 256x256. Since depth maps are an essential part of the experiments they are generated for the purpose of this work with the following procedure. To normalize the dataset, pixel values of each image are divided by 255 (maximum pixel value). Images are then binarized using the threshold of 0.5. The resulting black and white images are extruded with a fixed depth which is proportional to their class value, as depicted in Figure~\ref{figure:2}. Furthermore, the dataset is manipulated through two mechanisms: dataset size (number of examples in training set) and noise addition to images.

\subsection{NYU Depth V2}
\begin{figure}[htbp]
  \centering
  \includegraphics[scale=0.40]{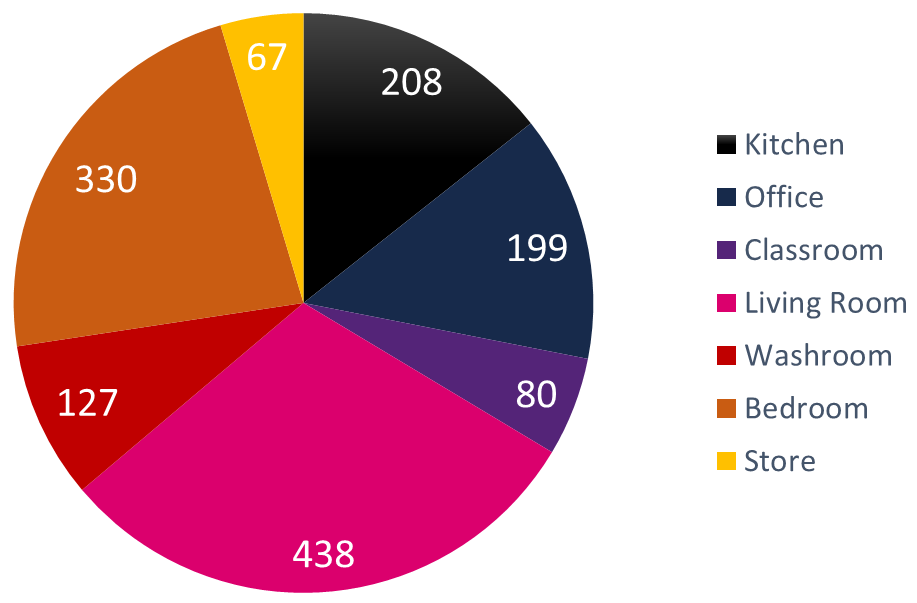}
  \caption{Number of images in each defined categories of the NYU-Depth-V2 dataset}
  \label{figure:3}
\end{figure}

\begin{figure}[htbp]
  \centering
  \includegraphics[scale=0.30]{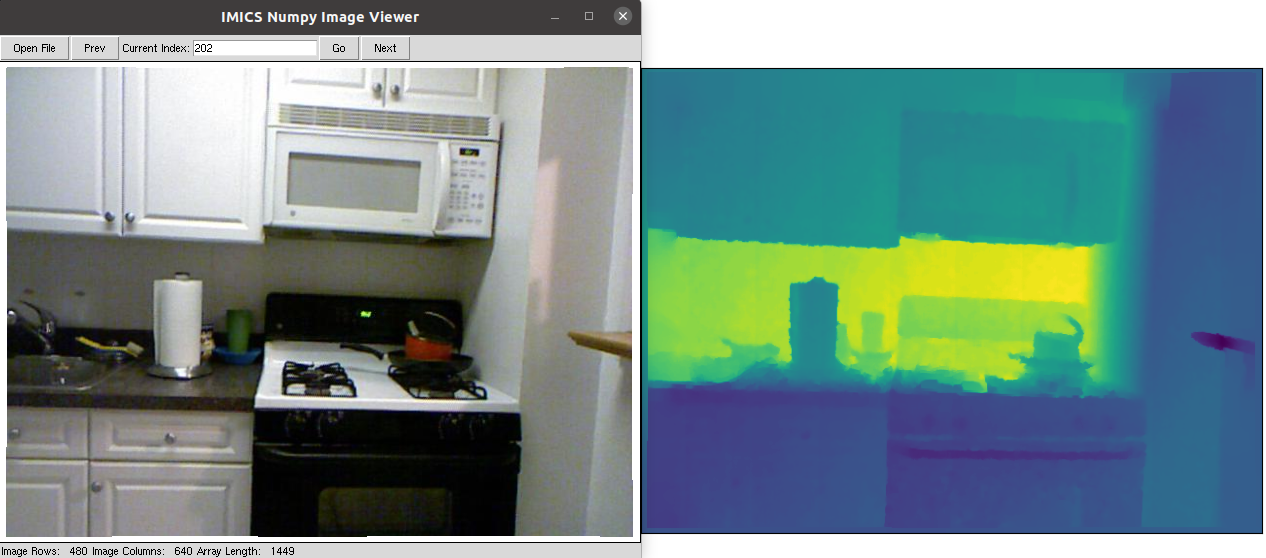}
  \caption{An example from NYU-Depth-V2 dataset illustrated with IMICS Numpy viewer}
  \label{figure:4}
\end{figure}

NYU-Depth-V2 ~\cite{Silberman:ECCV12} dataset includes RGB and depth images of indoor scenes, along with segmentation and classification of objects inside each image. The dataset is published in native formats of MATLAB and includes 1449 entries. As one the contributions of this research, we have preprocessed NYU-Depth-V2 and made it public in Python format\footnote{\url{https://www.dropbox.com/sh/xxbkzhrwpg6adrf/AACVZTjI3lBG89NXL3ySUrjBa?dl=0}}. The NYU dataset can be a good choice for a semantic segmentation task. However, the defined classification labels classify objects into structural classes that reflect their physical role in the scene, thus they are not appropriate for conventional classification purposes such as scene classification. Therefore, we manually labeled all images in the dataset for the scene classification task such that they are all divided into 7 categories: kitchen, office, classroom, living room, washroom, bedroom and store. Figure~\ref{figure:3} indicates the number of images in each category. We developed IMICS Numpy viewer\footnote{\url{https://github.com/IMICSLab/Numpy_Image_Viewer}} and made it publicly available as an open-source tool to facilitate the labeling process. Figure~\ref{figure:4} illustrates an image from NYU-Depth-V2 along with the corresponding depth map.

\section{Model Architecture and Multitask Configuration}
The main goal of this research is training a network for the depth regression and classification tasks to share parameters and enforce the learning of a common representation. Our network is given an image of a scene and produces two different outputs: depth map and class label. Depth maps contain the pixel-level depth information of the image, while class labels include the predicted class that the image belongs to. The fact that we aim to output depth maps implies the Fully Convolutional Networks (FCNs) are appropriate candidates for our purpose. Therefore, we opt U-Net \cite{Ronneberger0U-Net:}, which is the core of state-of-the-art approaches for segmentation tasks and benchmark of depth estimation \cite{Cho2019Large}, to act as the backbone of our model. In order to produce the classification output, we utilize the feature flow at the bottleneck of our U-Net.

\begin{figure}[htbp]
  \centering
  \includegraphics[scale=0.50]{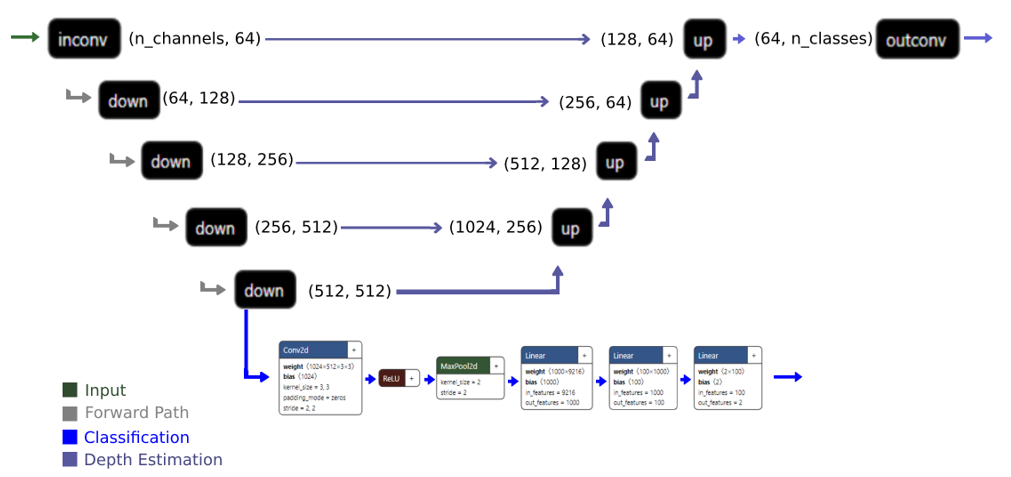}
  \caption{Model Architecture}
  \label{figure:5}
\end{figure}

\begin{figure}[htbp]
  \centering
  \includegraphics[scale=0.7]{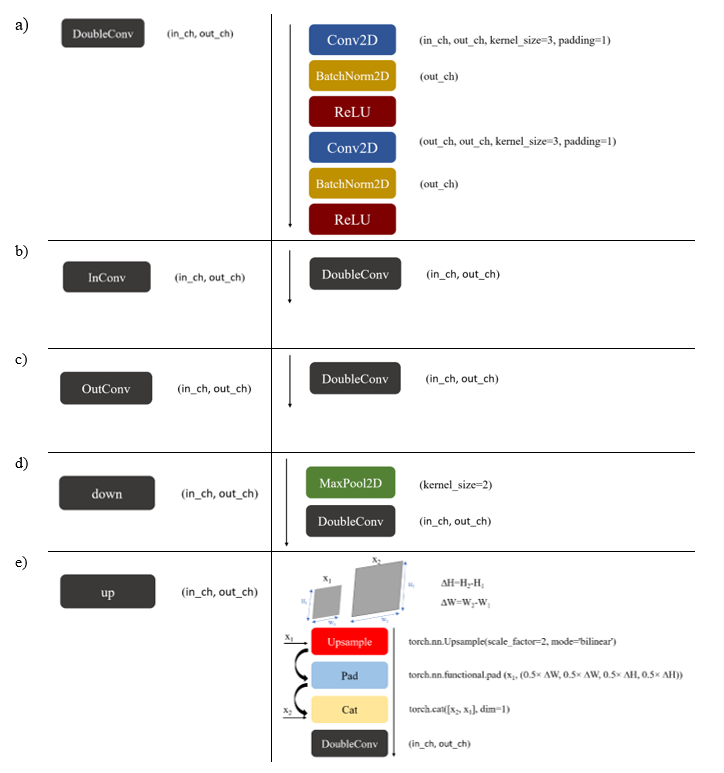}
  \caption{Model components}
  \label{figure:6}
\end{figure}

Our network simply generalizes U-Net to joint depth regression and classification tasks. U-net outputs a single depth map while we add a parallel branch that extracts features for the classification task and outputs the class label. The U-Net is composed of two sub-networks: encoding and decoding networks. The encoding network includes encoding blocks to learn input representations, and down sampling blocks to encode the input image at multiple levels into feature representations. In our experiments, this part is shared between the depth estimator and the classifier. The goal of the decoding network is to restore features learned by the encoder onto a higher-dimensional space similar to the original image. In our architecture, this part is depth-estimator-specific. U-Net also contains skip connections to enable encoder and decoder to share information. The parallel branch we add to the U-Net structure for classification purpose, uses the same encoder structure of the depth estimator U-Net, while in second stage, it adds a branch with fully connected layers to output classification labels. The full model architecture and details of each block of the network are indicated in Figures~\ref{figure:5} and~\ref{figure:6}.

There are multiple approaches to train a deep model with multitask learning. The simplest approach is sequential training where the model is first trained for the first task, then for the next tasks, in turn. In sequential training, hyperparameters may be shared among the tasks or a distinct set of hyperparameters can be defined for each task. The latter option has a higher potential, but it makes the grid search more complex. Another method is utilizing multiple optimizers. In sequential training, skills of the model on the first task usually degrade as it is being trained on the second task. Using multiple optimizers, the model is tuned for all the tasks at the same time and the ultimate performance across the tasks tends not to have a high variance. However, the training process may be unstable if the global optimums across the tasks are located in different parts of the parameter space. The last available approach is defining a multitask loss function which is the most popular option in literature. Multitask loss functions often have distinct components each corresponding to a task. If the multitask loss function is not defined in a proper manner, the optimization procedure might be confounded by a specific component of the loss function. Hence, this type of multitask training is vulnerable to under-fitting. 

To investigate the efficacy of our dMTL model, first, we used MNST dataset with different noise levels and training dataset sizes. For these experiments, we chose multiple optimizers approach for the loss function. Next, in a separate set of experiments using maximum noise level with minimum training dataset size for MNIST, we explored all methods for defining the loss function. This yielded multitask loss approach as the most optimal method, which we used for our experiments with NYU Depth dataset.

\section{Results and Discussion}
\subsection{MNIST Dataset}
Our experiments consist of three main parts. First, different amounts of noise were added to the input image and the accuracy of classification was calculated. Then we created images similar to depth maps for each image and used them to train a depth estimator along with our classification using dMTL approach. This approach robustly improved the classification performance. Next, the number of training images was manipulated and same approach as above was tested on each new dataset. Finally, we examined effect of different dMTL training methods to determine the best optimization method.

Throughout our experiments with MNIST we used a fixed set of hyperparameters. Our learning rate was \( 5\times10^{-5} \), L2 penalty was 0.001, momentum was chosen to be 0.9, batch size was equal to 4, we limited our experiments to 10 epochs, and we used weighted Cross Entropy (CE) as our loss function for the classification task. Given the fact that our dataset was almost balanced, unweighted CE would create same results. We used the first N examples of the MNIST training cohort for training. 1500 examples (index 45000 to 46500) from the MNIST training cohort formed our validation set and the first 1000 examples of the MNIST test set were used as our test cohort.

\begin{figure}[htbp]
  \centering
  \includegraphics[scale=0.45]{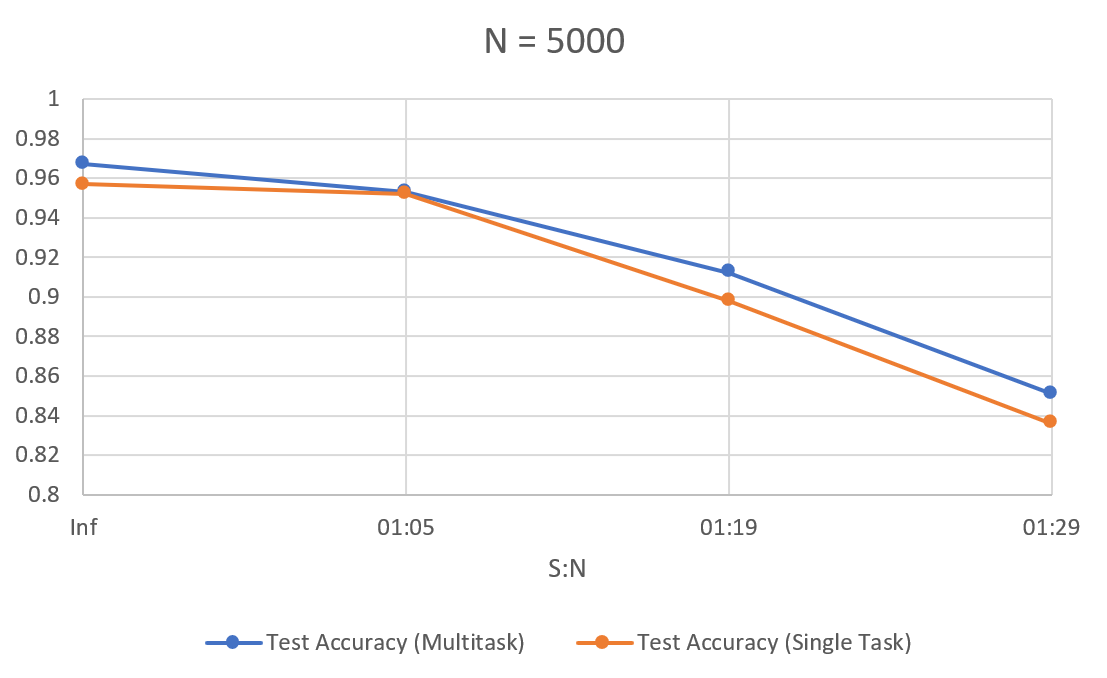}
  \caption{Image classification accuracy on 5000 training images with different noise levels using single and multitask learning methods (MNIST dataset)}
  \label{figure:7}
\end{figure}

Table~\ref{table:1} and Figure~\ref{figure:7} show results of experiments using 5000 training examples. As anticipated, higher levels of noise associate with lower performance of the model for both depth estimation and classification tasks. The model maintains most of its test accuracy in highly noisy environments, but depth estimation results are degraded. Although dMTL elevates classification performance at different noise levels, the test accuracy improvements are not considerable (e.g., ~2\% for SNR of 1:29).

\begin{table}
  \caption{Performance of the depth estimation model at different SNRs, with N=5000, with dMTL (MNIST dataset)}
  \label{table:1}
  \centering
  \resizebox{\textwidth}{!}{\begin{tabular}{llllll}
    S:N&	N&	Test Accuracy&	Input&	Depth Estimation Output&	Ground Truth Depth Map\\
    \midrule
    Inf&	5000&	0.967&  
    \begin{minipage}{.3\textwidth}
      \includegraphics[width=\linewidth, height=\imgdim, width=\imgdim]{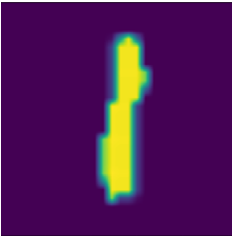}
    \end{minipage}&
    \begin{minipage}{.3\textwidth}
      \includegraphics[width=\linewidth, height=\imgdim, width=\imgdim]{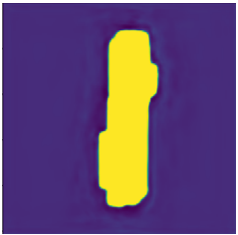}
    \end{minipage}&
    \begin{minipage}{.3\textwidth}
      \includegraphics[width=\linewidth, height=\imgdim, width=\imgdim]{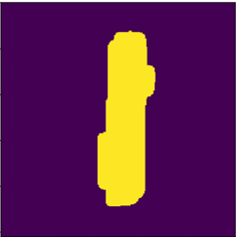}
    \end{minipage}\\
    \midrule
    1:5&	5000&	0.953&
    \begin{minipage}{.3\textwidth}
      \includegraphics[width=\linewidth, height=\imgdim, width=\imgdim]{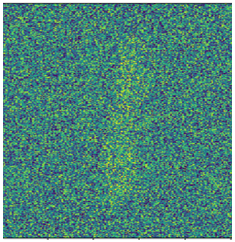}
    \end{minipage}&
    \begin{minipage}{.3\textwidth}
      \includegraphics[width=\linewidth, height=\imgdim, width=\imgdim]{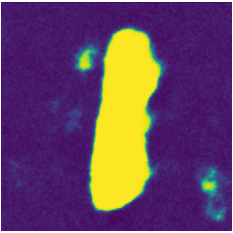}
    \end{minipage}&
    \begin{minipage}{.3\textwidth}
      \includegraphics[width=\linewidth, height=\imgdim, width=\imgdim]{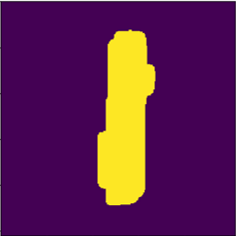}
    \end{minipage}\\
    \midrule
    1:19&	5000&	0.912&
    \begin{minipage}{.3\textwidth}
      \includegraphics[width=\linewidth, height=\imgdim, width=\imgdim]{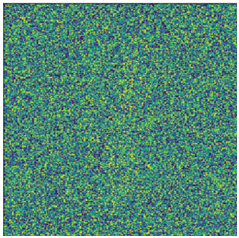}
    \end{minipage}&
    \begin{minipage}{.3\textwidth}
      \includegraphics[width=\linewidth, height=\imgdim, width=\imgdim]{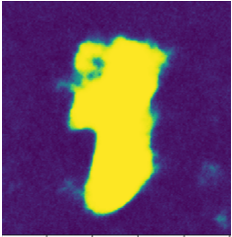}
    \end{minipage}&
    \begin{minipage}{.3\textwidth}
      \includegraphics[width=\linewidth, height=\imgdim, width=\imgdim]{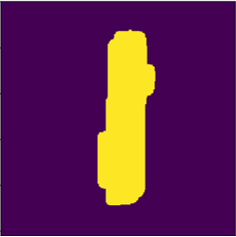}
    \end{minipage}\\
    \midrule
    1:29&	5000&	0.851&
    \begin{minipage}{.3\textwidth}
      \includegraphics[width=\linewidth, height=\imgdim, width=\imgdim]{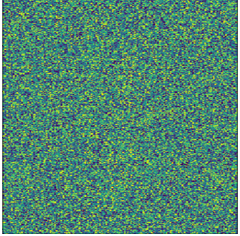}
    \end{minipage}&
    \begin{minipage}{.3\textwidth}
      \includegraphics[width=\linewidth, height=\imgdim, width=\imgdim]{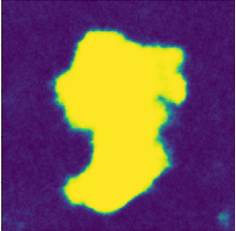}
    \end{minipage}&
    \begin{minipage}{.3\textwidth}
      \includegraphics[width=\linewidth, height=\imgdim, width=\imgdim]{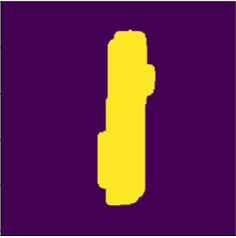}
    \end{minipage}\\
    \bottomrule
  \end{tabular}}
\end{table}

\begin{figure}[htbp]
  \centering
  \includegraphics[scale=0.45]{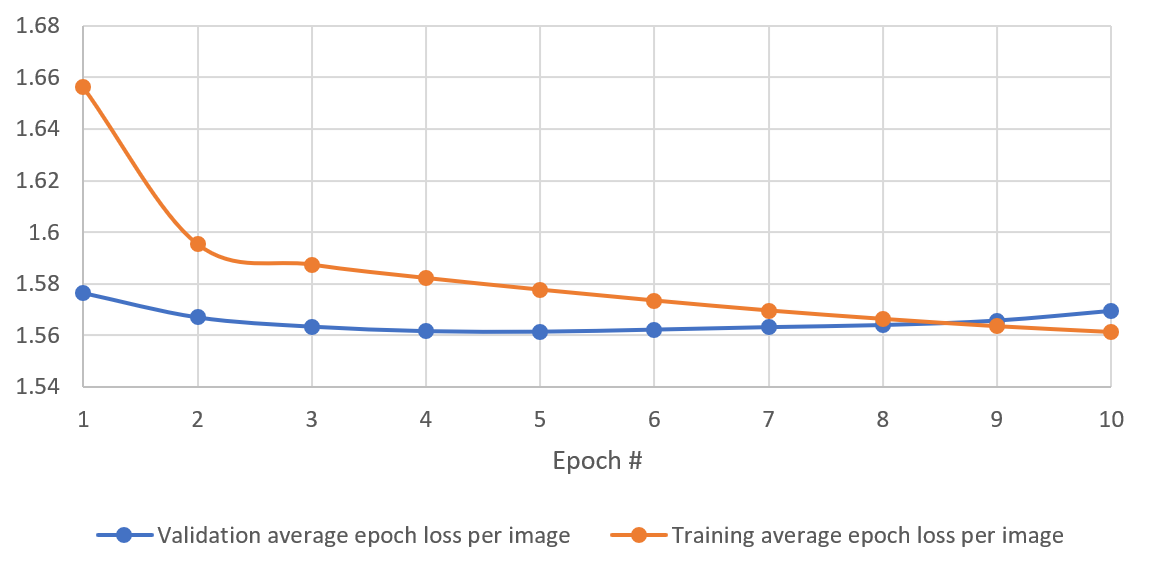}
  \caption{Performance of the model on the depth estimation task (MNIST dataset, N=5000, SNR=1:29, Multitask)}
  \label{figure:8}
\end{figure}

\begin{figure}[htbp]
  \centering
  \includegraphics[scale=0.45]{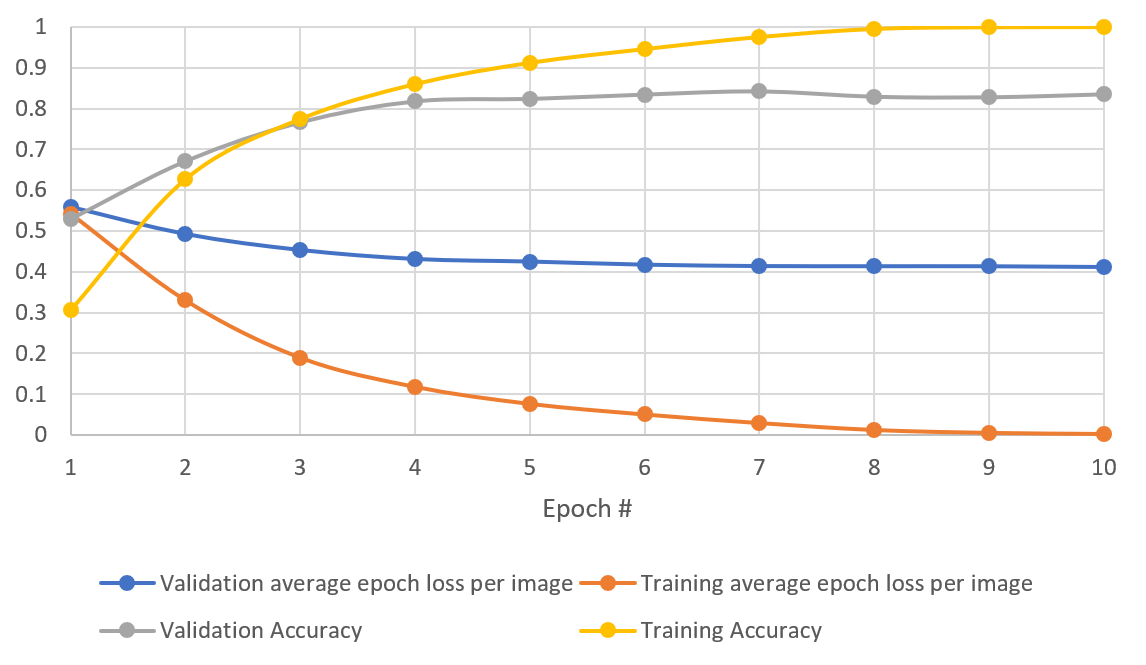}
  \caption{Performance of the model on the classification task (MNIST dataset, N=5000, SNR=1:29, Multitask)}
  \label{figure:9}
\end{figure}

\begin{figure}[htbp]
  \centering
  \includegraphics[scale=0.45]{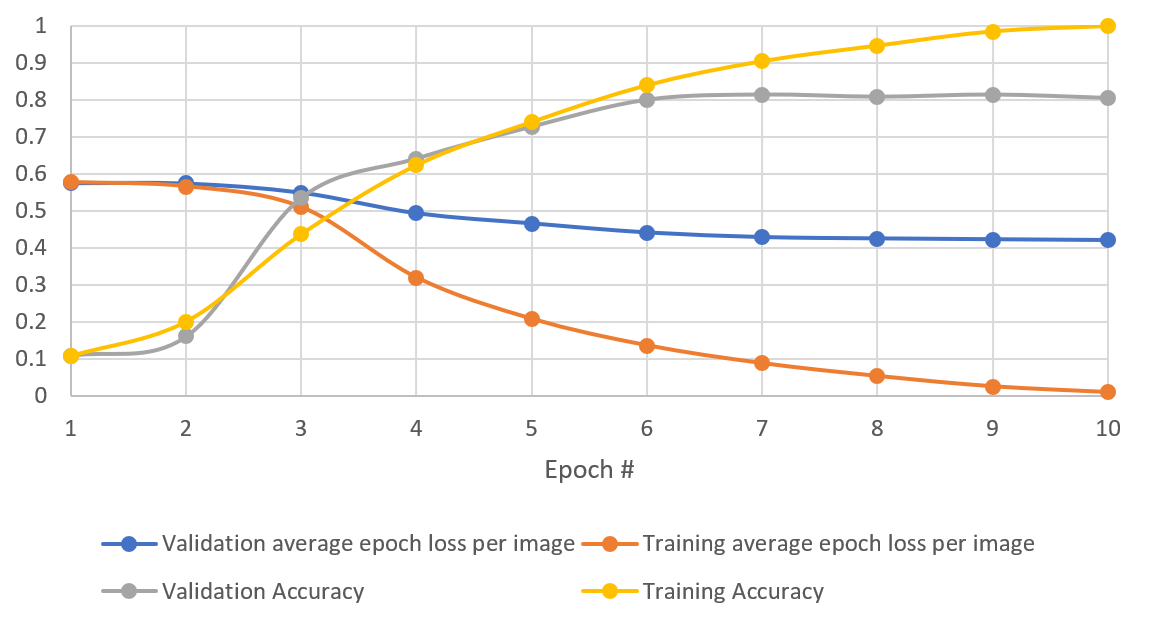}
  \caption{Performance of the model on the classification task (MNIST dataset, N=5000, SNR=1:29, Single Task)}
  \label{figure:10}
\end{figure}

Figures~\ref{figure:8},~\ref{figure:9}, and~\ref{figure:10} show details of the training process for the noisiest experiment (SNR=1:29) with 5000 training examples. As can be seen, if depth estimation is used as an auxiliary task, model convergence for the classification task is accelerated (accuracy of $\sim$0.80 is achieved in epoch 3 for dMTL vs epoch 6 for single task model). 

\begin{figure}[htbp]
  \centering
  \includegraphics[scale=0.50]{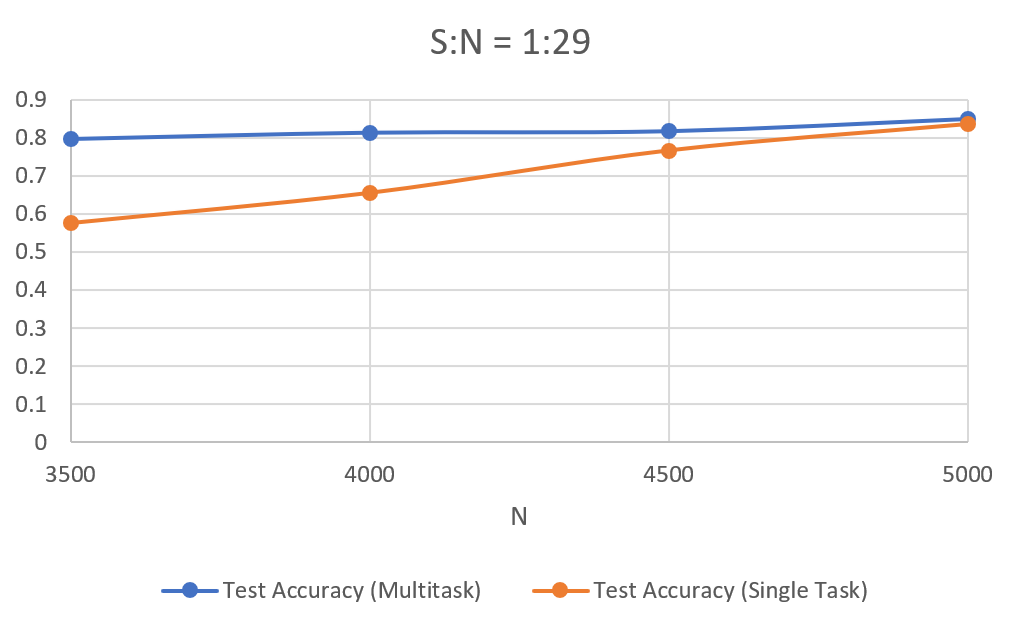}
  \caption{Image classification accuracy on various number of training images with 1:29 noise level using single and multitask learning methods (MNIST dataset)}
  \label{figure:11}
\end{figure}

Table~\ref{table:2} and Figure~\ref{figure:11} illustrate performance of the model at highest level of noise (SNR=1:29) with different training examples. The results demonstrate that if the number of training examples are limited, the effect of dMTL becomes bolder. For example, as it is seen in Figure~\ref{figure:11}, with data size of 3500, the dMTL model improves the test accuracy by more than 0.20 (0.6 to 0.8). These experiments illustrate that noise level has the greatest impact on depth estimation. However, classification is most affected by the size of the dataset.

In a separate set of experiments, we explored all the available options for loss function for our extreme case with the highest amount of noise and lowest number of examples (SNR = 1:29 and N = 3500). Table~\ref{table:3} contains the results. As it can be seen, since our focus is improving the classification accuracy and our auxiliary task, depth estimation, is used as a helper, the multitask loss approach achieves the most promising results in our experiments (accuracy of 0.875). Thus, we used multitask loss approach for our experiments with NYU Depth dataset.

\begin{table}
  \caption{Table 2: Performance of the depth estimation model at SNR = 1:29, with different number of training examples, with dMTL (MNIST dataset)}
  \label{table:2}
  \centering
  \resizebox{\textwidth}{!}{\begin{tabular}{llllll}
    S:N&	N&	Test Accuracy&	Input&	Depth Estimation Output&	Ground Truth Depth Map\\
    \midrule
    1:29&	5000&	0.851&
    \begin{minipage}{.3\textwidth}
      \includegraphics[width=\linewidth, height=\imgdim, width=\imgdim]{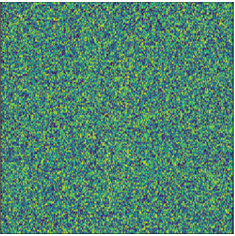}
    \end{minipage}&
    \begin{minipage}{.3\textwidth}
      \includegraphics[width=\linewidth, height=\imgdim, width=\imgdim]{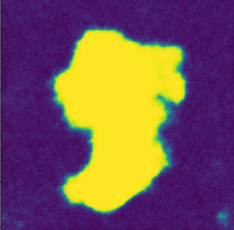}
    \end{minipage}&
    \begin{minipage}{.3\textwidth}
      \includegraphics[width=\linewidth, height=\imgdim, width=\imgdim]{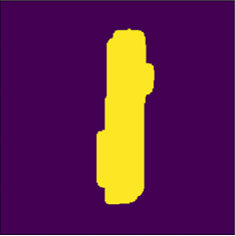}
    \end{minipage}\\
    \midrule
    1:29&	4500&	0.818&
    \begin{minipage}{.3\textwidth}
      \includegraphics[width=\linewidth, height=\imgdim, width=\imgdim]{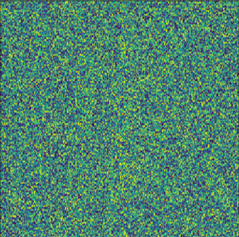}
    \end{minipage}&
    \begin{minipage}{.3\textwidth}
      \includegraphics[width=\linewidth, height=\imgdim, width=\imgdim]{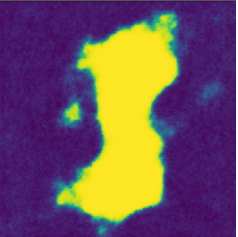}
    \end{minipage}&
    \begin{minipage}{.3\textwidth}
      \includegraphics[width=\linewidth, height=\imgdim, width=\imgdim]{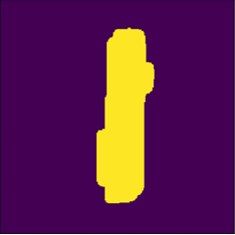}
    \end{minipage}\\
    \midrule
    1:29&	4000&	0.814&
    \begin{minipage}{.3\textwidth}
      \includegraphics[width=\linewidth, height=\imgdim, width=\imgdim]{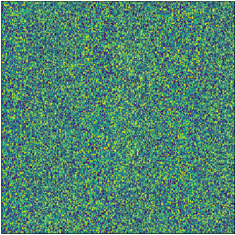}
    \end{minipage}&
    \begin{minipage}{.3\textwidth}
      \includegraphics[width=\linewidth, height=\imgdim, width=\imgdim]{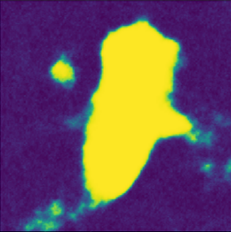}
    \end{minipage}&
    \begin{minipage}{.3\textwidth}
      \includegraphics[width=\linewidth, height=\imgdim, width=\imgdim]{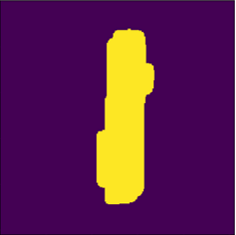}
    \end{minipage}\\
    \midrule
    1:29&	3500&	0.797&
    \begin{minipage}{.3\textwidth}
      \includegraphics[width=\linewidth, height=\imgdim, width=\imgdim]{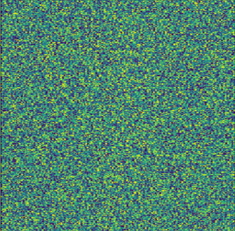}
    \end{minipage}&
    \begin{minipage}{.3\textwidth}
      \includegraphics[width=\linewidth, height=\imgdim, width=\imgdim]{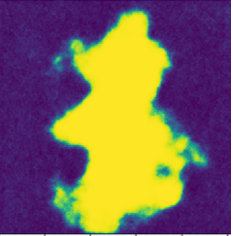}
    \end{minipage}&
    \begin{minipage}{.3\textwidth}
      \includegraphics[width=\linewidth, height=\imgdim, width=\imgdim]{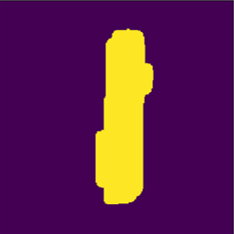}
    \end{minipage}\\
    \bottomrule
  \end{tabular}}
\end{table}

\begin{table}
  \caption{Performance of the model trained with different dMTL approaches (MNIST dataset)}
  \label{table:3}
  \centering
  \begin{tabular}{ll}
    Training Strategy&	Test Accuracy\\
    \midrule
    Single Task&	0.577\\
    \midrule
    Multitask, Sequential Training&	0.797\\
    \midrule
    Multitask, Multiple Optimizers&	0.858\\
    \midrule
    Multitask, Multitask-loss function&	0.875\\
    \bottomrule
  \end{tabular}
\end{table}

\subsection{NYU Depth V2 Dataset}
Due to the limited dataset size and uneven number of examples in each class, the NYU classifiers were incapable of separating all the seven classes with an acceptable accuracy (above 40\%). We visualized the dataset using the t-distributed stochastic neighbor embedding (tSNE)\cite{vanDerMaaten2008}, as illustrated in Figure~\ref{figure:11}, and did not observe any separation in the data. We believe the common elements in the scenes cause similarity between the scenes. For instance, chairs, desks, and computers are typical in classrooms and offices. Living rooms and bedrooms also share elements such as shelves and seats. Thus, we reformatted the scenario to a binary classification and chose Kitchen as the base class. In addition to simplifying the classification task, the approach lightened the computational costs of the experiments.

We applied what was learned from the MNIST experiments to the NYU dataset. As mentioned, the size of the dataset was limited and there was no room or necessity for manipulating the number of examples. Additionally, the overlapping objects in the scene classes acted as noise in the labels. Hence, adding noise to the input images would overcomplicate the classification task. Finally, we limited the setting to single task and multitasking with a multitask loss function. The results of the NYU experiments are shown in Table~\ref{table:4}. The experiments show that dMTL improves accuracy of the classifiers both on the validation and the test cohorts. 

\begin{figure}[htbp]
  \centering
  \includegraphics[scale=0.3]{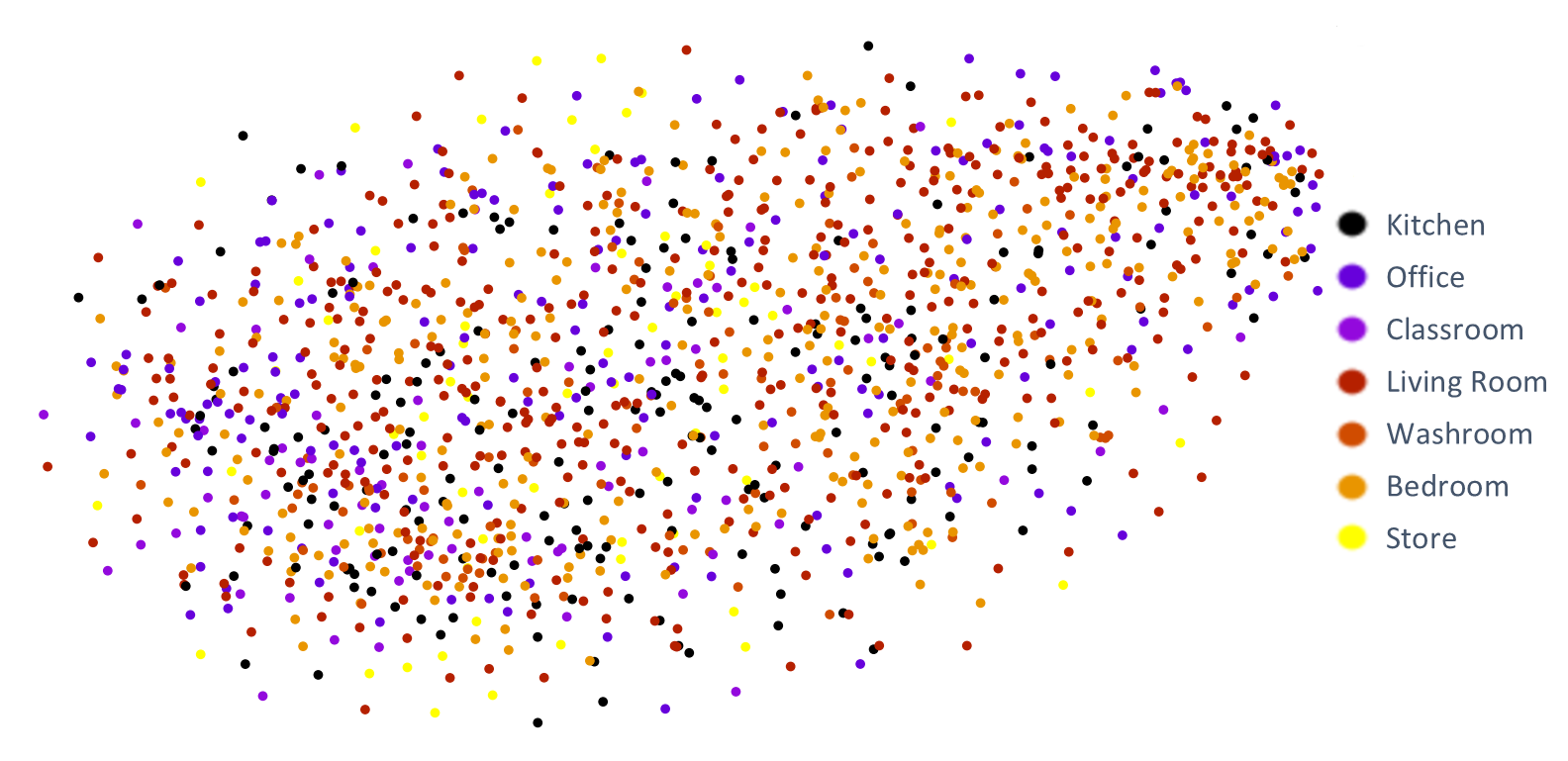}
  \caption{tSNE visualization of the NYU-Depth-V2 dataset with respect to the scene classification labels}
  \label{figure:12}
\end{figure}

\begin{table}
  \caption{NYU scene classification results}
  \label{table:4}
  \centering
  \begin{tabular}{lll}
    Cohort&	Average Accuracy (single task)& Average Accuracy(multitask)\\
    \midrule
    Validation&	0.767& 0.790\\
    \midrule
    Test&	0.756& 0.770\\
    \bottomrule
  \end{tabular}
\end{table}

\section{Conclusion}

In this paper, we presented a deep multitask learning (dMTL) method to improve the performance of the primary task. Our dMTL model improves the performance of the main task (i.e., image classification) in the presence of noise and limited data size, compared to a single task model. We optimized the effect of the auxiliary representations by training the data for the task of depth prediction, which benefits the main image classification task. In extensive experiments, our method was applied to the MNIST and NYU-Depth-V2 datasets and we showed how classification and depth estimation tasks can act as complementary operations and provide the network with more comprehensive information resulting in a better understanding of the images and improved classification performance. Furthermore, different loss function methods for the dMLT model were evaluated and it was revealed that multitask loss is the most effective method in a dMTL framework, achieving highest classification accuracy.

\begin{ack}
This research has been supported by Huawei Technologies Canada Co., Ltd.
\end{ack}

\printbibliography
\end{document}